\documentclass[letterpaper, 10 pt, conference]{ieeeconf}  

\IEEEoverridecommandlockouts                              

\usepackage{graphicx} 
\usepackage{amsmath} 
\usepackage{hyperref}  

\title{\LARGE \bf
Toward Low-Flying Autonomous MAV Trail Navigation using Deep Neural Networks for Environmental Awareness
}

\author{Nikolai Smolyanskiy \and Alexey Kamenev \and Jeffrey Smith \and Stan Birchfield
\thanks{The authors are affiliated with NVIDIA Corporation, Redmond, WA 98052, USA.  Email:   
{\tt\small \{nsmolyanskiy, akamenev, jeffreys, sbirchfield\}@nvidia.com}}
}

\begin{document}

\newcommand{\bfx}{\ensuremath{{\bf x}}}
\newcommand{\bfy}{\ensuremath{{\bf y}}}
\newcommand{\id}[1]{\ensuremath{\mathop{\mathit{#1}}\nolimits}}

\maketitle
\thispagestyle{empty}
\pagestyle{empty}

\begin{abstract}
We present a micro aerial vehicle (MAV) system, built with inexpensive off-the-shelf hardware, for autonomously following trails in unstructured, outdoor environments such as forests.  The system introduces a deep neural network (DNN) called TrailNet for estimating the view orientation and lateral offset of the MAV with respect to the trail center.  The DNN-based controller achieves stable flight without oscillations by avoiding overconfident behavior through a loss function that includes both label smoothing and entropy reward.  In addition to the TrailNet DNN, the system also utilizes vision modules for environmental awareness, including another DNN for object detection and a visual odometry component for estimating depth for the purpose of low-level obstacle detection.  All vision systems run in real time on board the MAV via a Jetson TX1.  We provide details on the hardware and software used, as well as implementation details.  We present experiments showing the ability of our system to navigate forest trails more robustly than previous techniques, including autonomous flights of 1~km.

\end{abstract}

\section{Introduction}

Autonomously following a man-made trail in the forest is a challenging problem for robotic systems.  Applications for such a capability include, among others, search-and-rescue, environmental mapping, wilderness monitoring, and personal videography.  In contrast to ground-based vehicles, micro aerial vehicles (MAVs) offer a number of advantages for solving this problem: they are not limited by the difficulty or traversability of the terrain, they are capable of much higher speeds, and they have the ability to quickly switch from one trail to another by flying through or over the forest.  

In order for a complete MAV system to follow a trail, it must not only detect the trail in order to determine its steering commands, but it also must be aware of its surroundings.  An MAV that lacks such a capability is in danger of colliding with overhanging branches or, even worse, with people or pets using the trail.  Environmental awareness is therefore a critical component for trail-following robots, particularly for low-flying MAVs.

In this paper, we present a MAV system for autonomous trail following. The system uses a deep neural network (DNN) that we call TrailNet for determining the MAV's view orientation and lateral offset within the trail. The computed pose is then used for continuous control to allow the MAV to fly over forest trails. In addition, vision modules for environmental awareness enable the MAV to detect and avoid people and pets on the trail, as well as to estimate depth in front of the robot for the purpose of reactively avoiding obstacles.  All subsystems described in this paper run simultaneously in real time on board the MAV using an NVIDIA Jetson TX1 computer.  On-board processing is essential for ensuring the safety of such a mission-critical system.

In this work, we make the following contributions:
\begin{itemize}
\item	A hardware/software system for environmentally aware autonomous trail navigation using DNNs that runs in real time on board an MAV.  This is the first such system of which we are aware that can navigate a narrow forested trail of 1 km autonomously.\footnote{\url{https://www.youtube.com/watch?v=H7Ym3DMSGms}, \url{https://www.youtube.com/watch?v=USYlt9t0lZY}}
\item	A novel DNN architecture for trail detection with improved accuracy and computational efficiency via a less-confident classification scheme for more stable control as well as additional categories for estimating both view orientation and lateral offset.
\item	A methodology for retraining a DNN with 6 categories (for view orientation and lateral offset) by transfer learning from a network with 3 categories (orientation only).
\end{itemize}

\section{Previous Work}

Today, commercial MAVs are mostly teleoperated, use GPS for navigation, and have limited ability to automatically avoid obstacles. Nevertheless, the research community has made significant strides toward introducing more visual-based processing to facilitate obstacle avoidance and state estimation.  Alvarez~et~al.~\cite{alvarez2016} use a structure-from-motion (SfM) algorithm to estimate depth, and hence avoid obstacles, from a single forward-facing camera.  Bachrach~et~al.~\cite{bachrach2012} build a map using an RGBD camera, which is then used for localization and planning.  Bry~et~al.~\cite{bry2012} combine an inertial measurement unit (IMU) with a laser range finder to localize a fixed-wing MAV, enabling reliable flight in indoor environments.  Faessler~et~al.~\cite{faessler2016} apply the SVO algorithm \cite{forster2014} to a downward-facing camera and IMU to build a map of the environment, which is then used for planning.  Fraundorfer~et~al.~\cite{fraundorfer2012} fuse a forward-facing stereo pair of cameras with a downward-facing camera for map building and state estimation.  Scaramuzza~et~al.~\cite{scaramuzza2014ram} also combine IMU and three cameras for obstacle avoidance and map building.  Some of the challenges of flying low among obstacles are described by Scherer~et~al.~\cite{scherer2008ijrr}.

Recent attention has also been paid toward using deep reinforcement learning (DRL) for MAVs.  In foundational work, Bhatti~et~al.~\cite{bhatti2016} describe a system that combines visual mapping with visual recognition to learn to play first-person video games using DRL.  Similarly, Lillicrap~et~al.~\cite{lillicrap2016} apply DRL to learn end-to-end policies for many classic continuous physics problems.  Ross~et~al.~\cite{ross2013icra} apply DRL to learn to avoid obstacles in the context of an MAV flying in a forest.  Sadeghi and Levine~\cite{sadeghi2016arx} apply DRL to enable an MAV to learn to fly indoors after training only on synthetic images.  Zhang~et~al.~\cite{zhang2015arx} use DRL to learn obstacle avoidance policies for a simulated quadrotor.

Our work is partly inspired by, and is most closely related to, the trail DNN research of Giusti~et~al.~\cite{giusti2016}, who trained a convolutional DNN that predicts view orientation (left/right/straight) by collecting an extensive dataset via a head-mounted rig consisting of three cameras.  The trained DNN was then used on board an MAV to determine steering commands to follow a trail.  This work builds on the earlier ground-based robotic work of Rasmussen~et~al.~\cite{rasmussen2014fsr}.

We were also inspired by NVIDIA's DNN-controlled self-driving car~\cite{bojarski2016}, which uses three cameras located on the car's dashboard to collect training data for three different directions. During training, the system computes correct steering effort for each virtual view (created by interpolating footage between the real cameras) to keep the vehicle straight and in the lane.  In this manner, the DNN learns to associate visual input with the correct steering command.  The trained DNN then steers the car using a single camera, on a variety of terrains and environments.  We adapted this approach by augmenting the IDSIA forest trail dataset~\cite{giusti2016} with our own footage collected with a three-camera wide-baseline rig to collect side views. In this manner, our system is able to estimate both lateral offset and view orientation within the trail.

\section{System Description}

\begin{figure}
	\centering
		\includegraphics[width=\columnwidth]{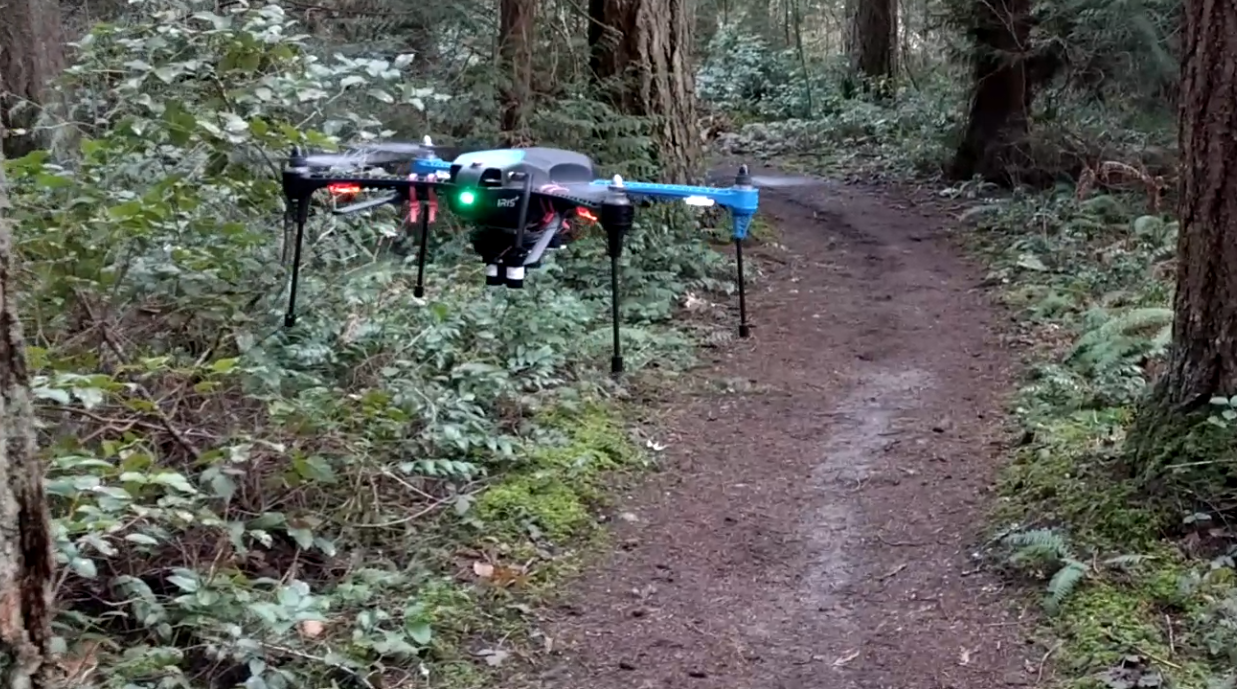}
	\caption{Our MAV following a trail in a forest.}
	\label{fig:droneOnTrail}
\end{figure}

To ensure a robust and flexible platform for forest flight experiments, we chose to use inexpensive off-the-shelf components.  Our hardware setup, shown in Figure~\ref{fig:droneOnTrail}, includes a 3DR Iris+ quadcopter with open source Pixhawk autopilot, and NVIDIA Jetson TX1 with Auvidea J120 carrier board.  The vision processing described in this paper uses a single forward-facing Microsoft HD Lifecam HD5000 USB camera, run in 720p mode at 30 fps.  All processing is done on the Jetson TX1.  The 3DR Iris+ quadcopter could be replaced by a similar 450--550 drone, such as a DJI Flame Wheel, if desired.

The MAV is equipped with a downward-facing, high framerate (400 Hz) PX4FLOW optical flow sensor with sonar, and Lidar Lite V3.  Developed for the PX4 / Pixhawk project by ETH~\cite{honegger2013}, PX4FLOW provides reliable metric position and attitude estimation by computing 2D optical flow on the ground, then using ground distance measurements and gyro readings to compute vehicle ground speed. Once computed, this speed is sent to an extended Kalman filter (EKF) running on Pixhawk to be fused with other sensor measurements (e.g., IMU) for even more precise state estimation. 

\begin{figure}
	\centering
		\includegraphics[width=\columnwidth]{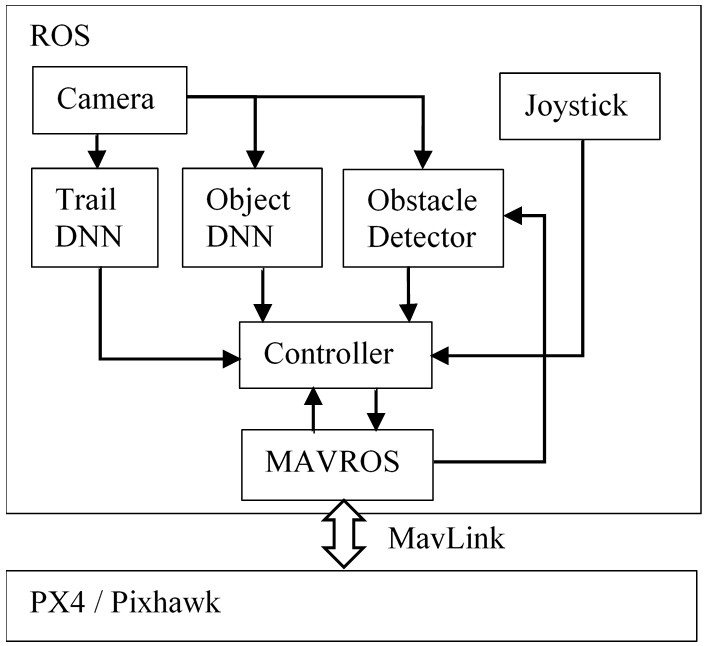}
	\caption{System diagram.}
	\label{fig:architecture}
\end{figure}

The diagram in Figure~\ref{fig:architecture} illustrates our software architecture.  We adopted the open source PX4 flight stack~\cite{meier2011icra} as firmware for the Pixhawk autopilot.  PX4 provides a flexible and robust control over a range of MAV configurations.  It includes software-in-the-loop (SITL) simulation, which we use for controller testing and debugging. The Jetson TX1 on-board computer communicates with PX4 via MavLink protocol.
We run the Robot Operating System (ROS) and Ubuntu Linux4Tegra (L4T) on the Jetson TX1.  As shown in the figure, the system uses the following ROS nodes:  GSCam for reading USB camera input, JOY for reading Xbox/Shield game controller commands used for teleoperation (during training and for emergency override), and MAVROS for communicating with PX4.

\begin{figure*}
	\centering
		\includegraphics[width=2\columnwidth]{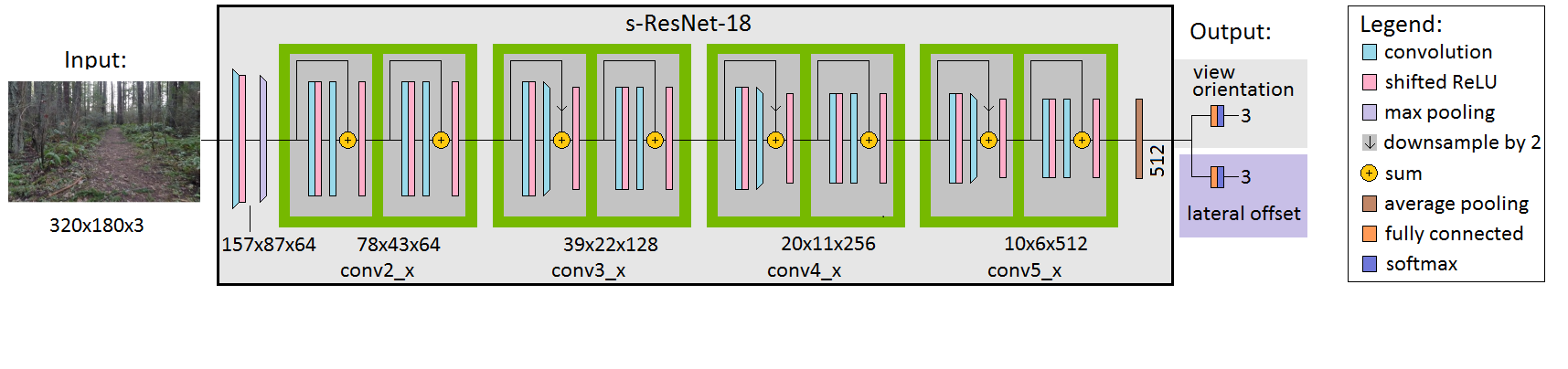}
	\caption{TrailNet architecture used in this paper to determine the MAV's view orientation and lateral offset with respect to the trail center.  The bulk of the network (labeled s-ResNet-18) is the standard ResNet-18~\cite{he2016} architecture, but without batch normalization and with ReLU replaced by shifted ReLU.  The first convolution layer is $7 \times 7$, whereas all the others are $3 \times 3$.  Some layers downsample using stride of 2; all others use stride of 1.  All weights except the lateral offset layer are trained with the IDSIA trail dataset, after which these weights are frozen, and the lateral offset layer weights are trained with our dataset.  A final joint optimization step could be performed but was deemed to be not necessary.}
	\label{fig:trailnetarch}
\end{figure*}

Vision processing is performed by three nodes.  A DNN node using TensorRT runtime applies our TrailNet trained by
Caffe/DIGITS.  An object detection node runs a version of the YOLO DNN \cite{redmon2016arx}.  And a third node runs the DSO visual odometry algorithm \cite{engel2017}, whose output is converted to a camera-centric depth map for obstacle detection and avoidance.  

The controller node is responsible for computing desired movement commands (waypoints) per current trail DNN predictions, detected obstacles/objects and teleoperation commands.  For safety, the teleoperation commands always take precedence over DNN predictions, so a human operator can override the MAV at any time to prevent undesirable movements. In our final experiments, this overriding was not neccessary. The computed waypoint is then sent to MAVROS which resubmits it to PX4 via MavLink. We use a right-handed ENU (east-north-up) inertial coordinate frame for waypoint computation, which must be converted to PX4's right-handed NED (north-east-down) coordinate frame. The controller node runs at 20 Hz.

A few additional system details are worth mentioning.  We added a Lidar Lite V3 laser range finder to improve the distance-to-ground estimation. We also replaced PX4FLOW's default 16 mm narrow-angle lens with a 6 mm wide-angle lens, changed the firmware to accommodate the new focal length, and used custom PX4FLOW firmware that provides better exposure control. These changes considerably improved the system's performance in low light environments such as forests and in situations when flying debris blown by the MAV's rotors reduces optical flow accuracy.


\section{Vision-Based Processing}

In this section we describe the processing performed by the three vision-based modules using the forward-facing camera.

\subsection{Estimating lateral offset and view orientation with DNN}

Figure~\ref{fig:trailnetarch} shows the TrailNet DNN used in this work for estimating the MAV's lateral offset and view orientation with respect to the trail.  The network is standard ResNet-18~\cite{he2016} but with several changes:  we removed batch normalization; replaced ReLUs with shifted ReLU (SReLU)~\cite{clevert2016} activation functions, which are computationally efficient approximations to the highly effective exponential linear units (ELUs) \cite{clevert2016}; and introduced a double-headed fully connected output layer.  


Like \cite{giusti2016}, we treat the problem as one of classification with soft labels, rather than regression, because of the ease of data collection and labeling, as well as reduced sensitivity to noise.  A significant difference with~\cite{giusti2016} is that our network outputs 6 categories rather than 3 to capture not only the 3 orientations with respect to the trail (facing left / facing center / facing right) but also the 3 lateral offsets (shifted left / centered / shifted right) with respect to the trail center.  In our experience, these additional categories are essential to accurate state estimation and hence to reliable trail following.  With just 3 categories, if the MAV flies close to the trail edge but parallel to the trail, the DNN will not cause it to turn, because it is already oriented straight. Eventually, this negligence will cause it to hit tree branches near the edge.  Thus, robust navigation is precluded with just 3 categories, because of the difficulty of resolving the ambiguity of needing to turn to correct an orientation error, and needing to fly in a non-straight path to correct a lateral offset error.

Training DNNs requires large amounts of data.  Our approach to training the 6-category DNN applies principles of transfer learning.  We first trained a 3-category DNN to predict view orientation (left / center / right), see Figure~\ref{fig:trailnetarch}, using the IDSIA Swiss Alps trail dataset available from~\cite{giusti2016}.  
This dataset includes footage of hikes recorded by 3 cameras (aimed left 30$^\circ$, straight, and right 30$^\circ$), with enough cross-seasonal and landscape variety to train all but the lateral offset layers of our network.  

\begin{figure}
	\centering
		\includegraphics[width=0.6\columnwidth]{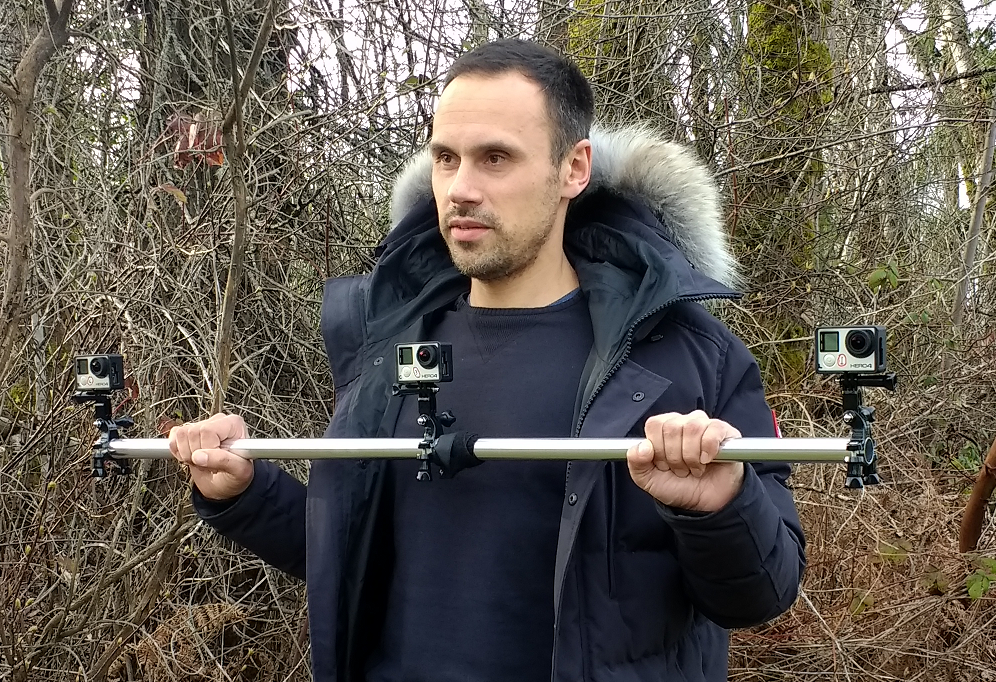}
	\caption{Three-camera wide-baseline rig used to gather training data in the Pacific Northwest.  Note that, unlike the rig used in~\cite{giusti2016}, the lateral separation between cameras in this rig allows the network to determine the lateral offset of the MAV.  The baseline between the left and right cameras is 1~m, and the center camera is within 0.5~cm of the midpoint.}
	\label{fig:rig}
\end{figure}

To train the rest of the network, we used a dataset of our own collected in the Pacific Northwest using the 3-camera rig shown in Figure~\ref{fig:rig}, in which GoPro cameras (120$^\circ$ $\times$ 90$^\circ$ FOV) were mounted on a 1 meter bar (so the distance between adjacent cameras was 0.5~m). Utilizing a Segway MiniPro, we gathered data over several trails by recording video footage of all cameras simultaneously as the Segway was driven along the middle of the trail.  The videos were cropped to 60$^\circ$ horizontal FOV and used to train the second 3-category head of the network used to predict lateral offset (left / center / right), while freezing all the other weights.\footnote{Originally we created, from each of the 3 physical cameras, 3 virtual 60$^\circ$ horizontal FOV cameras aimed left, straight, and right.  This yielded 9 simultaneous videos covering the 3 lateral positions and 3 orientations for each trail.  Although these data could be used to train the full 6-category network, we found it sufficient to use only the 3 central views to train the lateral offset layer, and moreover, the approach outlined here better demonstrates the network's ability to generalize across different locales.}  Note that, unlike the rig used in~\cite{giusti2016}, the wide baseline between our cameras (similar to the work of~\cite{bojarski2016}) enables the network to disambiguate lateral offset from view orientation.

During training, we use a typical data augmentation pipeline: random horizontal flips (with appropriate label changes); random contrast (0.2 radius), brightness (0.2 radius), saturation (0.4 radius), sharpness (0.3 radius) and jitter with random permutations of these transformations; random scale jitter (min: 0.9, max: 1.2); random rotations (max abs angle: 15); and random crops.  

We experimented with different loss functions.  The most robust results were obtained using the following loss function:
\begin{equation}
{\cal L} = \underbrace{-\sum_i p_i \ln(y_i)}_{\text{cross entropy}}-\underbrace{\lambda_1(-\sum_i y_i \ln(y_i))}_{\text{entropy reward}} + \underbrace{\lambda_2 \phi(\bfy)}_{\text{side swap} \atop \text{penalty}}
\label{eq:loss_function}
\end{equation}
where $p_i$ and $y_i$ are the smoothed ground truth label and network prediction of category $i \in \{\id{left},\id{center},\id{right}\}$, respectively, $\bfy$ is a 3-element vector containing all $y_i$, $\lambda_1$ and $\lambda_2$ are scalar weights, and the side swap penalty penalizes gross mistakes (i.e., swapping left and right):
\begin{equation}
\phi(\bfy) = 
  \begin{cases}
	y_{\id{left}} & \text{if } {\hat i} = {\id{right}} \\
	y_{\id{right}} & \text{if } {\hat i} = {\id{left}} \\	
	0 & \text{if } {\hat i} = {\id{center}} 
	\end{cases}
\end{equation}
where ${\hat i} = \arg \max_i p_i$ is the ground truth category.  This loss function is used for training both the view orientation and lateral offset heads of the network.\footnote{The side swap penalty was only used for the lateral offset head.  For the view orientation head, we set $\phi(\bfy)=0$ because it was not needed.}  Label smoothing \cite{szegedy2015arx} in the first term and the entropy reward term \cite{mnih2016arx} together help reduce overconfident behavior of the network, which proved to be critical to stable flight.  

We created a novel controller to mix probabilistic DNN predictions. This design simplifies transforming DNN predictions into flight commands. Each time our trail DNN sends its prediction, the controller computes a small counterclockwise turning angle $\alpha$ by
\begin{equation}
	\alpha = \beta_1 (y_{\id{right}}^{\id{vo}} - y_{\id{left}}^{\id{vo}}) + \beta_2 (y_{\id{right}}^{\id{\ell o}} - y_{\id{left}}^{\id{\ell o}}),  \label{eq:control}
\end{equation}
which is the weighted sum of differences between the right and left predictions of the view orientation ($\id{vo}$) and lateral offset ($\id{\ell o}$) heads of the network, where $\beta_1$ and $\beta_2$ are scalar angle parameters that control turning speed.  We set $\beta_1=\beta_2=10^\circ$. Once the turn angle is known, we compute a new waypoint located some distance in front of the MAV and at a turn angle relative to MAV's body frame. The distance to the waypoint affects the vehicle's speed. We set the orientation to face the new waypoint direction. The waypoint is converted to the inertial frame and sent (with a fixed distance to the ground) to MAVROS module for PX4 flight plan execution.  

\subsection{Object detection with DNN}

For safety, it is important for the system not to collide with objects such as people, cars, animals, and so on.  The object detection DNN is responsible for detecting such objects, which may not be visible to the SLAM system (described in the next section) if they are independently moving. For each detected object, the network provides bounding box image coordinates.  When the box size exceeds a threshold percentage of the image frame, the controls from Equation~\eqref{eq:control} are overridden to cause the MAV to stop and hover.  We explored several popular object detection DNN architectures:
\begin{itemize}
	\item R-CNN with "fast" variants~\cite{girshick2015}.  Unfortunately, these are not a viable option to run in real time on the Jetson TX1.
  \item 	SSD~\cite{liu2016eccv} is a single shot, fast, and accurate network.  However, it uses custom layers which are not currently supported by TensorRT and therefore cannot run efficiently on Jetson TX1.
  \item 	YOLO~\cite{redmon2015arx} (and the more recent version~\cite{redmon2016arx}) is single shot, fast, and accurate. The original YOLO DNN uses some features (like leaky ReLU activations) which are not currently supported by TensorRT, or features like batch normalization which are not supported in 16-bit floating-point mode (FP16).
\end{itemize}

\begin{figure}
	\centering
		\includegraphics[width=\columnwidth]{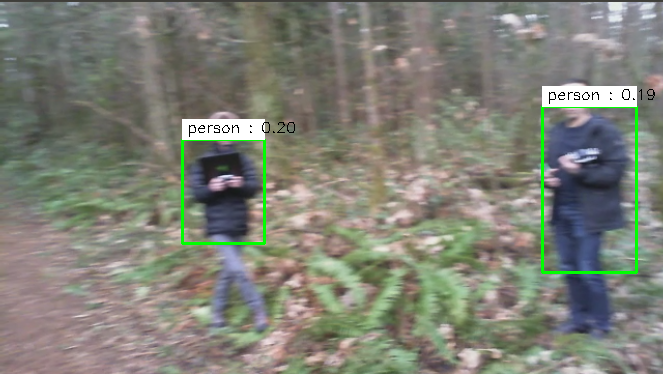}
	\caption{People on the trail detected by YOLO running on board the MAV.}
	\label{fig:yolo}
\end{figure}

As a result, the network that we have chosen to use for object detection is based on YOLO~\cite{redmon2015arx} with small modifications to allow efficient execution using FP16 mode of TensorRT on Jetson TX1. We retrained the network on the PASCAL VOC dataset, which is the same dataset used in the original YOLO.  An example of the YOLO output on an image from the camera is shown in Figure~\ref{fig:yolo}.

\subsection{Obstacle detection and avoidance with Monocular SLAM}

In addition to higher-level navigation tasks such as trail following and semantic object detection, an autonomous MAV must be able to quickly detect and avoid obstacles in its path, even if these obstacles have never been seen before (and hence not available for training). Such obstacles can be static, such as trees or walls, or dynamic, such as people, animals, or moving vehicles.  For safety reasons, the system should be robust in the presence of complex, highly dynamic environments such as forest trails, and in the face of changes in lighting, shadows, and pose.

One approach to low-level obstacle detection is traditional stereo-based matching.  However, motivated by the desire to develop a system that could eventually be miniaturized, as well as to minimize system complexity, we focused on monocular techniques.  A number of depth-from-single-image techniques have been developed over the years using both traditional machine learning~\cite{michels20015icml,saxena2008pami} as well as more recent deep learning methods~\cite{eigen2014,liu2015pami}.  Other approaches have been developed that estimate depth using a monocular video stream~\cite{pizzoli2014icra,ummenhofer2016arx}.

\begin{figure}
	\centering
		\includegraphics[width=\columnwidth]{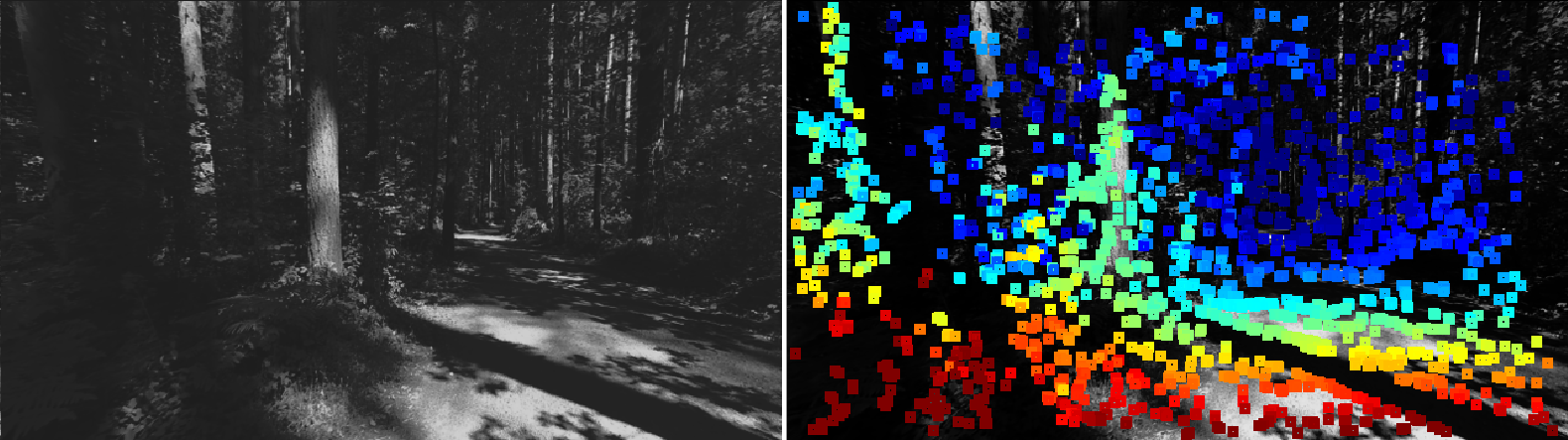}
	\caption{An image (left) from the on-board camera, and the sparse depth map (right) from DSO, using a jet color map (red is close, blue is far).}
	\label{fig:dso_both}
\end{figure}

For this work, we chose to use the direct sparse odometry (DSO) method described in~\cite{engel2017}.  DSO computes semi-dense 3D maps of the environment and is computationally efficient.  We used the open-source version of this code, which we adapted to run efficiently on the ARM architecture of the Jetson TX1.  On this platform, DSO provides 30 Hz updates to the camera pose estimates as well as the depth image, and updates its keyframes at between 2 and 4 Hz.  Figure~\ref{fig:dso_both} shows an example image from the forward-facing camera, along with a pseudocolored depth map computed by DSO.

During flight, DSO provides an estimate of the camera's global pose $\bfx_{dso}$ and an inverse depth image $I_{dso}$. Using the rectified camera intrinsics, we convert $I_{dso}$ into a 3D point cloud in the local space of the camera. To use these points for navigation or obstacle detection, we must put them in the MAV's frame of reference, including proper metric scale.  This similarity transform is determined using the odometry information provided by the MAV's other sensors.

\begin{figure}
	\centering
		\includegraphics[width=\columnwidth]{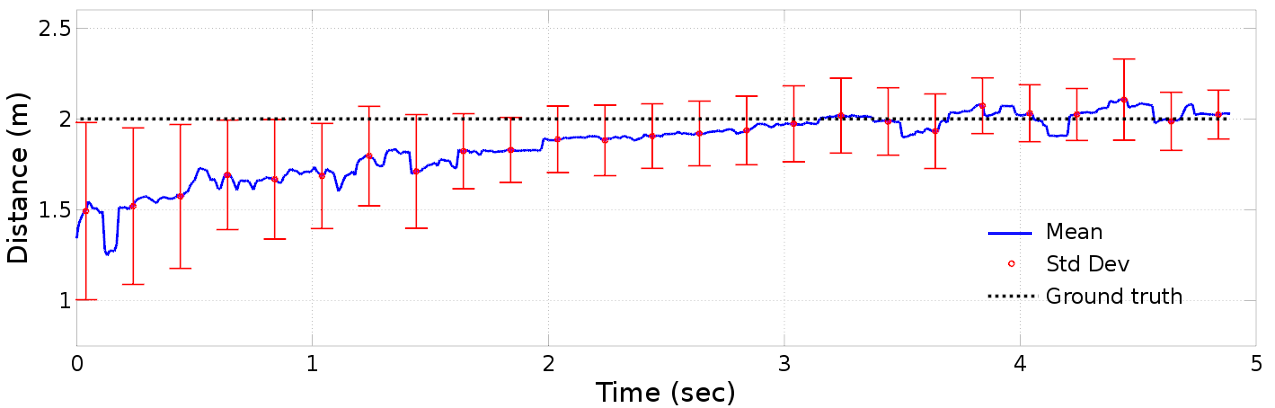}
	\caption{Estimating absolute distance by aligning the DSO and MAVROS poses using Procrustes analysis.  After several seconds, the estimation settles to the ground truth value.  Shown are the mean and $\pm 1 \sigma$ from five trials.}
	\label{fig:distance_estimation}
\end{figure}

Specifically, at any given time PX4FLOW + Pixhawk is used to generate an estimate of the current pose ($\bfx_{mav}$) of the vehicle in the inertial frame.  When the visual odometry system provides an $\bfx_{dso}$, $I_{dso}$ pair, we associate it with the latest pose.  Once several of these measurement pairs have been generated, Procrustes analysis is used to generate the rigid similarity transformation between the MAV's frame of reference and that of the visual odometry measurements.  This transform is then used to put the points extracted from $I_{dso}$ into the MAV's frame of reference, where we can then detect nearby objects. The set of measurement pairs ($\bfx_{dso}$ paired with $\bfx_{mav}$) is updated when the MAV's motion generates sufficient parallax. These pairs are kept in a circular buffer, so eventually older measurements are discarded in favor of new ones.  This moving window has the effect of potentially changing the metric scale over time, but more importantly it allows the system to be robust to measurement error and avoid pathological initializations.  An experiment showing metric depth estimation is illustrated in Figure~\ref{fig:distance_estimation}.

%

\section{Experimental Results}

%

\subsection{DNN comparison}

In developing our trail detection network, we explored several architectures: ResNet-18~\cite{he2016}, SqueezeNet~\cite{iandola2016}, a miniatured version\footnote{To fit onto the TX1, several layers of AlexNet were removed.} of AlexNet~\cite{krizhevsky2012}, the DNN architecture of Giusti et al.~\cite{giusti2016},\footnote{Our implementation of the Giusti et al. DNN replaces $\tanh$ with ReLU, which significantly improves training speed with no loss in accuracy.} and our final TrailNet architecture.\footnote{In early experiments we also tried VGGNet-16~\cite{simonyan2015iclr}, but the network was too large to fit onto the TX1.}  After much experimentation, we settled on the proposed architecture due primarily to its ability to achieve reliable autonomous flight over long distances.  Other important criteria in comparing architectures included accuracy in category predictions, computational efficiency, and power efficiency.  

Table~I compares the different network architectures.  All architectures output 6 categories (that is, 3 probabilities for view orientation and 3 probabilities for lateral offset), except for Giusti et al.~\cite{giusti2016}, which only outputs 3 categories.  For quantitative comparison, we used the IDSIA trail dataset~\cite{giusti2016} collected on Swiss Alps forest trails. Since details on dataset preparation are not provided in~\cite{giusti2016}, we have chosen to use trails 001, 002, 004, 007, 009 for training and 003, 008 and 010 for testing. As the number of images is different in each trail, we used undersampling and oversampling to balance the dataset.  

Note that the classification accuracy on the dataset does not necessarily correlate with the closed-loop system's ability to navigate autonomously.  Autonomy was measured by the percentage of time that the MAV did not need to be overridden by a human operator to prevent collision with trees/bushes on either side of the trail as the MAV flew along a specific 250~m trail in the Pacific Northwest that was not encountered during our data collection.  This same trail was used to test each network.  Only our TrailNet was able to fly autonomously through the entire 250~m trail, despite that it did not achieve the highest classification accuracy on the offline dataset.


\begin{table*}
	\centering
		\begin{tabular}{c|c|c|c|c|c|c|c|c}
			DNN & $n_{cat}$ & loss terms & $n_{layer}$ & $n_{param}$ & $t_{train}$ & $t_{run}$ & accuracy & autonomy \\
			\hline
			Giusti et al. \cite{giusti2016} & 3 & YNN & 6 & 0.6M & 2 hr.\ & 2 ms & 79\% & 80\% \\
			ResNet-18 & 6 & YNN & 18 & 10M & 10 hr.\ & 19 ms$^*$ & 92\% & 88\% \\
			mini AlexNet & 6 & YNN & 7 & 28M & 4 hr.\ & 8 ms & 81\% & 97\% \\
			SqueezeNet & 6 & YYY & 19 & 1.2M & 8 hr.\ & 3 ms & 86\% & 98\% \\ 
			TrailNet (ours) & 6 & YYY & 18 & 10M & 13 hr.\ & 11 ms & 84\% & 100\%
			%
		\end{tabular}
	\caption{Comparison of various DNN architectures for trail following.  The columns show the number of categories ($n_{cat}$), whether the loss terms (CE/ER/SSP) from Equation~\eqref{eq:loss_function} are used, number of layers ($n_{layer}$), number of parameters ($n_{param}$), training time ($t_{train}$), per-frame execution time ($t_{run}$), accuracy on the IDSIA dataset, and autonomy on a 250~m trail.  Note that our TrailNet achieves the highest autonomy despite not yielding the highest accuracy.  {\tiny $^*$All models were run in FP16 mode, except ResNet-18, which was run in FP32 mode.}} 
	%
	\label{tab:DNNComparison}
\end{table*}


\subsection{Autonomous trail following}

\begin{figure}
	\centering
		\includegraphics[width=\columnwidth]{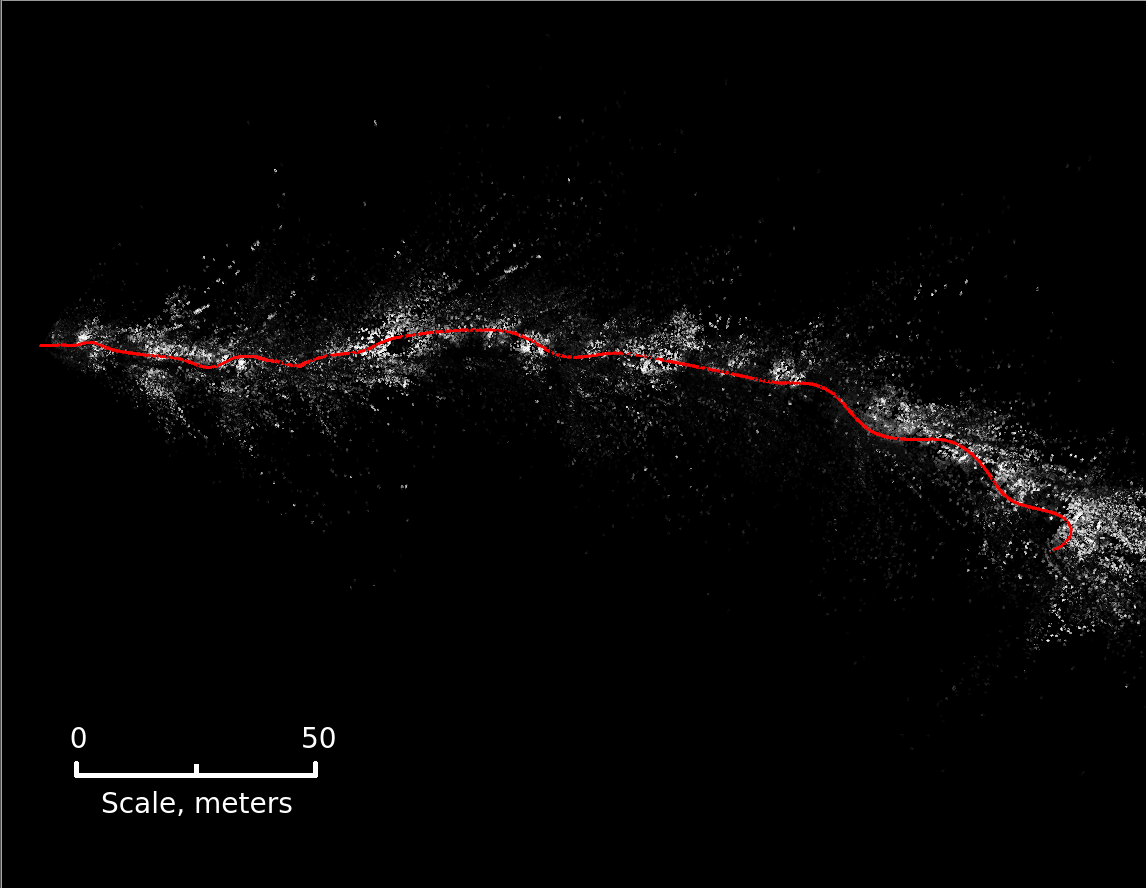}
	\caption{Top-down view of a 250 meters drone trajectory (red) through the forest, with 3D map overlaid (gray dots) generated by DSO SLAM.}
	\label{fig:path}
\end{figure}

We tested our TrailNet system by autonomously navigating in several different environments, including a 100~m fairly straight forest trail, a 250~m zigzag forest trail with 6 turns (the same trail used in the experiments above), a 1~km zigzag forest trail over hilly terrain, and a straight trail in an open field.  The latter trail was particularly difficult to detect due to lack of defining features with respect to the surrounding grass.  The trails were approximately 1.5 m wide, and the height ranged from 1 to 2~m.  Figure~\ref{fig:path} shows a trajectory through the zigzag trail.

One surprising discovery was that an overconfident network actually performs worse in practice than one that provides less confident predictions.  Our original network yielded 0.9--0.99 probability for the winning class, leaving very low probabilities for the remaining classes.  As a result, the MAV delayed turning until too late due to extreme bias toward flying straight.  Essentially, this behavior was caused by the network overfitting to the training data.  To prevent this overconfident behavior, we added the entropy reward term and label smoothing to our loss function, Equation~\eqref{eq:loss_function}, which makes our model less confident and allows the system to mix predicted probabilities to compute smoother trail direction angles for better control.  

\begin{figure}
	\centering
		\includegraphics[width=\columnwidth]{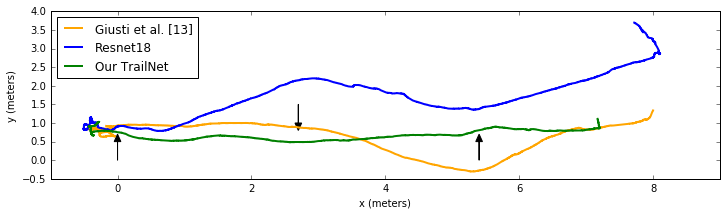}
	\caption{Top-down trajectories of MAV (traveling left to right) controlled by three different DNNs (3-class Giusti et al. DNN~\cite{giusti2016}, 3-class standard ResNet-18, and our 6-class TrailNet DNN).  The MAV was disturbed by a 2-second rotation at three locations (indicated by arrows).  Note that our network recovered from the disturbances faster than the other networks, following a nearly straight path.}
	\label{fig:trajectories}
\end{figure}

To further demonstrate the improved performance of our system, we designed an experiment in which the MAV was flown along a straight trajectory along a 2~m wide trail at 2~m/s speed at 2~m altitude.  We disturbed the trajectory by injecting a rotation to the left or right for 2 seconds, after which we let the MAV self-correct to fly straight in order to turn back to the trail.  Figure~\ref{fig:trajectories} displays the results of this experiment.  A 3-class view orientation-only standard ResNet-18 performed the worst:  The MAV simply followed the path created by the disturbances, with almost no noticeable correction.  The 3-class Giusti et al. DNN~\cite{giusti2016} also struggled to recover in a reasonable amount of time after each disturbance.  In contrast, our 6-class TrailNet DNN allowed the MAV to quickly return to the straight course after each disturbance, resulting in a nearly straight trajectory.

%








%

All modules run simultaneously on the Jetson TX1.  Together, these systems use 80\% of the CPU (4 cores) and 100\% of the GPU, when running DSO at 30~Hz, TrailNet at 30~Hz, and YOLO at 1~Hz.  Note that DSO and the DNNs are complementary to each other, since the former uses the CPU, while the latter use the GPU.


%
%

\section{Conclusion}

We have presented an autonomous MAV system that is capable of navigating forest trails under the canopy using deep neural networks (DNNs).  Our TrailNet DNN improves upon existing work in the following ways:  utilizing shifted ReLU activation functions for increased accuracy and computational efficiency, incorporating 3 additional categories via transfer learning to enable the estimation of both lateral offset and view orientation with respect to the trail, and a novel loss function to prevent overconfidence in the network.  Together, these improvements yield smoother and more robust control.  We also demonstrate the ability to generate environmental awareness with additional perceptual modules, including DNN-based object detection and obstacle detection via monocular visual odometry.  All software runs simultaneously in real time on a Jetson TX1 on board the MAV.  Together, these improvements take us one step closer toward fully autonomous, safe robotic navigation in unstructured environments. 

\section{Acknowledgment}

Many thanks to Simon Baker and Nicholas Haemel for supporting this project and to Jonathan Tremblay for helpful suggestions regarding the loss function.
  
\newpage

\end{document}